\documentclass[11pt]{article}
\usepackage{coling2018}
\usepackage{times}
\usepackage{latexsym}
\usepackage{url}
\usepackage{multirow}
\usepackage{graphicx}
\usepackage{amsmath}
\usepackage{amssymb}
\usepackage{makecell}
\usepackage{url}
\usepackage{tabularx}
\usepackage[hidelinks,pdftitle={Representation Learning of Entities and Documents from Knowledge Base Descriptions},pdfauthor={Ikuya Yamada, Hiroyuki Shindo, Yoshiyasu Takefuji}]{hyperref}

\title{Representation Learning of Entities and Documents from \\ Knowledge Base Descriptions}

\newcolumntype{C}[1]{>{\centering\let\newline\\\arraybackslash\hspace{0pt}}m{#1}}

\author{
  \begin{tabular}{C{3.3cm} C{3.5cm} C{4.5cm}}
    Ikuya Yamada\textsuperscript{1,3} & Hiroyuki Shindo\textsuperscript{2,3} & Yoshiyasu Takefuji\textsuperscript{4} \\
    {\tt \footnotesize{ikuya@ousia.jp}} & {\tt \footnotesize{shindo@is.naist.jp}} & {\tt \footnotesize{takefuji@sfc.keio.ac.jp}} \\
  \end{tabular}
  \\
  \\
  {$^1$ Studio Ousia},
  {$^2$ Nara Institute of Science and Technology},\\
  {$^3$ RIKEN AIP},
  {$^4$ Keio University}\\
}

\date{}

\begin{document}
\maketitle
  \begin{abstract}
    In this paper, we describe \textit{TextEnt}, a neural network model that learns distributed representations of entities and documents directly from a knowledge base (KB).
    Given a document in a KB consisting of words and entity annotations, we train our model to predict the entity that the document describes and map the document and its target entity close to each other in a continuous vector space.
    Our model is trained using a large number of documents extracted from Wikipedia.
    The performance of the proposed model is evaluated using two tasks, namely fine-grained entity typing and multiclass text classification.
    The results demonstrate that our model achieves state-of-the-art performance on both tasks.
    The code and the trained representations are made available online for further academic research.
  \end{abstract}

  \section{Introduction}

  \blfootnote{
      \hspace{-0.65cm}  
      This work is licensed under a Creative Commons
      Attribution 4.0 International License.
      License details:
      \url{http://creativecommons.org/licenses/by/4.0/}
  }

  The problem of learning distributed representations (or embeddings) from a knowledge base (KB) has recently attracted considerable attention.
  These representations enable us to use the large-scale, human-edited information of a KB in machine learning models, and can be applied in various natural language tasks such as entity linking \cite{hu-EtAl:2015:ACL-IJCNLP,Yamada2016,TACL1065}, entity search \cite{hu-EtAl:2015:ACL-IJCNLP}, and link prediction \cite{Bordes2013,wang-EtAl:2014:EMNLP20145}.

\begin{figure}[t]
    \centering
    \includegraphics[width=14cm,clip]{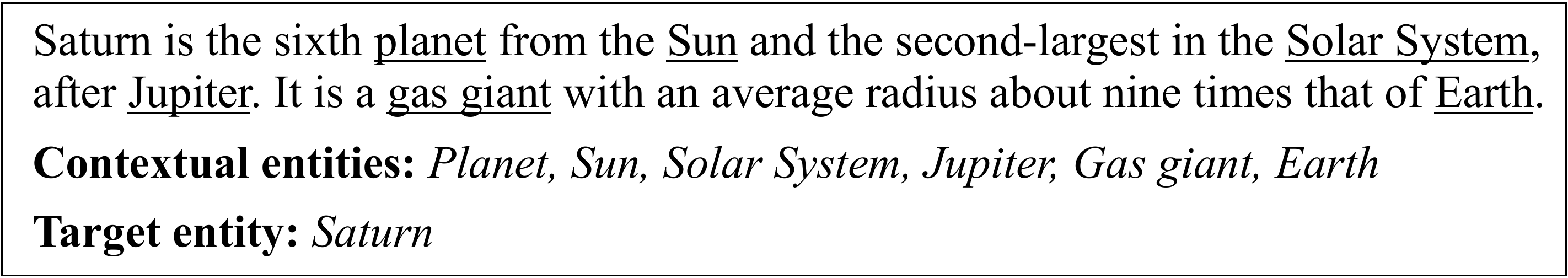}
    \label{fig:kb-description}
    \caption{Example of a KB document with entity annotations.}
  \end{figure}

  In this paper, we describe \textit{TextEnt}, a simple neural network model that learns distributed representations of entities and documents from a KB.
  Specifically, given a document in a KB consisting of words and \textit{contextual} entities (i.e., entities referred from entity annotations in the document), our model predicts the \textit{target} entity explained by the document (see Figure \ref{fig:kb-description}), and maps the document and its target entity close to each other in a continuous vector space.
  Here, words, contextual entities, and target entities are mapped into continuous vectors that are updated throughout the training.
  In this study, we train the model using documents retrieved from Wikipedia.

  One key characteristic of our model is that it enables us to combine the semantic signals obtained from both words and entities in a straightforward manner.
  The main motivation for using entities in addition to words is to address the problems of ambiguity (i.e., the same words or phrases may have different meanings) and variety (i.e., the same meaning may be expressed using different words or phrases) in natural language.
  For example, the word \textit{Washington} is ambiguous because it can refer to a US state, or the capital city of the US, or the first US president \textit{George Washington}, and so on.
  Further, \textit{New York} is sometimes referred to as \textit{NY} or by its nickname, the \textit{Big Apple}.
  Obviously, entities do not have these problems, because they are uniquely identified in the KB.

  To evaluate our model, we address two important tasks using the proposed representations.
  Firstly, we consider a fine-grained entity typing task \cite{yaghoobzadeh-schutze:2015:EMNLP} to evaluate the quality of the learned entity representations.
  In this task, the aim is to infer one or more types of each entity (e.g., \textit{athlete}, \textit{airport}, \textit{sports\_team}) from a predefined type set.
  We perform this task using the simple multilayer perceptron (MLP) classifier with the learned entity representations as features.
  The results show that our method outperforms the state-of-the-art methods by a wide margin.

  Secondly, we consider a multiclass text classification task, which aims to classify documents into a set of predefined classes.
  This task examines the ability of our model as a generic encoder of arbitrary documents.
  One important approach adopted here is that we automatically annotate entities appearing in the target documents using a publicly available entity linking system and encode the documents to the document representations in the same manner as the documents in the KB.
  For this task, the logistic regression classifier is applied to the document representations.
  Because of the quality of semantic signals obtained from the entities, our method outperforms strong state-of-the-art methods on two popular datasets (i.e., the 20 newsgroups dataset \cite{Lang1995} and R8 dataset \cite{Debole2005}).
  To facilitate further research, our code and the trained representations are available online at \url{https://github.com/studio-ousia/textent/}.

  Our contributions can be summarized as follows:

  \begin{itemize}
    \item We propose \textit{TextEnt}, a simple neural network model that learns distributed representations of entities and documents from a KB.
    Given a document in a KB consisting of words and contextual entities, our model learns the representations by predicting the target entity explained by the document (see Figure \ref{fig:kb-description}).
    We train our model using large-scale documents extracted from Wikipedia.
    \item Our proposed model allows us to effectively combine the semantic signals retrieved from both words and entities in a straightforward manner.
    We demonstrate the effectiveness of this feature by addressing two important tasks: fine-grained entity typing and text classification.
    Despite the simplicity of our approach, we achieve state-of-the-art results in both tasks.
    \item We have published our code and the trained representations online to facilitate further academic research.
  \end{itemize}

  \section{Our Method}

  In this section, we describe our approach of learning distributed representations of entities and documents from a KB.

  \subsection{Model}
  \label{subsec:model}
  \begin{figure}[t]
    \centering
    \includegraphics[width=8.5cm,clip]{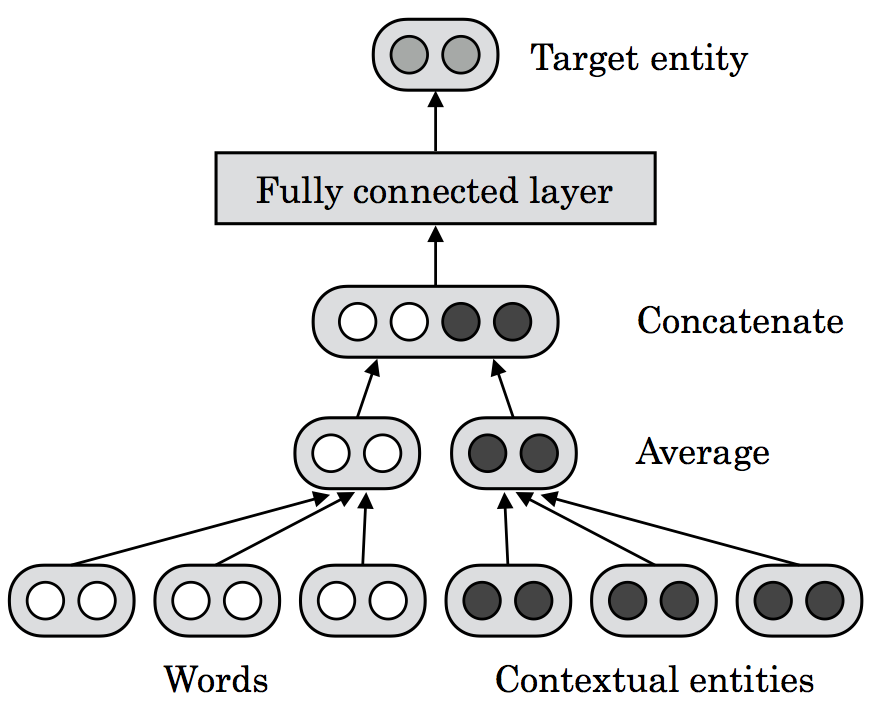}
    \caption{Model architecture of TextEnt.}
    \label{fig:architecture}
  \end{figure}

  Given a document $D$ in a KB consisting of a set of words $w_1, ..., w_N$ and a set of contextual entities $e_1, ..., e_K$, we train our model to predict the target entity that the document is explaining.
  We first derive two vector representations of document $D$: the word-based representation $\mathbf{v}_{D_w}$ and the contextual entity-based representation $\mathbf{v}_{D_e}$.
  For simplicity, we compute $\mathbf{v}_{D_w}$ and $\mathbf{v}_{D_e}$ by averaging the vector representations of words and those of contextual entities, respectively.
  \begin{equation}
  \mathbf{v}_{D_w} = \frac{1}{N}\sum_{n=1}^{N} \mathbf{a}_{w_n},\;
  \mathbf{v}_{D_e} = \frac{1}{K}\sum_{n=1}^{K} \mathbf{b}_{e_n},
  \label{eq:vector-averaging}
  \end{equation}
  where $\mathbf{a}_w \in \mathbb{R}^d$ and $\mathbf{b}_e \in \mathbb{R}^d$ are the vector representations of words and contextual entities, respectively.

  We define a probability that represents the likelihood of entity $e_t$ being the target entity of document $D$ as the following softmax function:
  \begin{equation}
  P(e_t|D) = \frac{\exp(\mathbf{c}_{e_t}\!^\top \mathbf{v}_D)}{\sum_{e' \in E_{KB}}\exp(\mathbf{c}_{e'}\!^\top \mathbf{v}_D)},
  \label{eq:softmax}
  \end{equation}
  where $E_{KB}$ is a set of all entities in the KB, $\mathbf{c}_{e} \in \mathbb{R}^d$ denotes the vector representation of target entity $e$, and $\mathbf{v}_D \in \mathbb{R}^d$ is the vector representation of document $D$.

  Here, $\mathbf{v}_D$ is computed using a fully connected hidden layer with $\mathbf{v}_{D_w}$ and $\mathbf{v}_{D_e}$ as inputs:
  \begin{equation}
  \mathbf{v}_D = \mathbf{W}[\mathbf{v}_{D_w}, \mathbf{v}_{D_e}]
  \label{eq:v_d}
  \end{equation}
  where $\mathbf{W} \in \mathbb{R}^{d\times 2d}$ is a weight matrix, and $[\mathbf{v}_i, \mathbf{v}_j]$ is the concatenation of $\mathbf{v}_i$ and $\mathbf{v}_j$.
  This layer projects the input vector ($[\mathbf{v}_{D_w}, \mathbf{v}_{D_e}]$) down to $d$ dimensions, and captures the interactions between $\mathbf{v}_{D_w}$ and $\mathbf{v}_{D_e}$.

  We use the categorical cross-entropy loss to train the model:
  \begin{equation}
  \mathcal{L} = - \sum_{(D, e_t) \in \Gamma} \log P(e_t|D),
  \label{eq:loss}
  \end{equation}
  where $\Gamma$ represents a set of pairs consisting of a document $D$ and its target entity $e_t$ in the KB.

  When training our model, the denominator in Eq. \eqref{eq:softmax} is computationally expensive because it involves summation over all KB entities.
  To address this, we use negative sampling \cite{Mikolov2013a};
  specifically, we replace $E_{KB}$ in Eq. \eqref{eq:softmax} with a set consisting of the target entity $e_t$ and $k$ randomly chosen negative entities.
  Furthermore, to avoid overfitting, we use \textit{word dropout} \cite{iyyer-EtAl:2015}, which randomly excludes words and contextual entities with a probability $p$ during the training.

  We also test models trained using only words (denoted by \textit{TextEnt-word}) and only contextual entities (denoted by \textit{TextEnt-entity}) in our experiments.
  These variants are created by replacing $\mathbf{v}_D$ in Eq. \eqref{eq:softmax} with $\mathbf{v}_{D_w}$ (\textit{TextEnt-word}) and $\mathbf{v}_{D_e}$ (\textit{TextEnt-entity}).
  Hereafter, our original model is referred to as \textit{TextEnt-full}.

  \subsection{Dataset}

  We trained our model using documents obtained from the April 2016 version of the DBpedia NIF abstract dataset\footnote{\url{http://wiki.dbpedia.org}}, which contains the texts and entity annotations in the first introductory sections of Wikipedia articles.

  For computational efficiency, we limited the size of our dataset.
  In particular, we excluded documents with fewer than five incoming links from other documents if the corresponding entity of the document is not contained in the dataset used in our fine-grained entity typing experiments, presented in Section \ref{subsec:entity-typing}.
  As a result, the number of target documents was 702,388.

  We also modified all words to lowercase, and excluded words that make fewer than five appearances and contextual entities that make fewer than three appearances in the documents.
  Thus, the final dataset contained 242,771 unique words and 327,263 unique contextual entities.

  \subsection{Parameters}
  \label{subsec:parameters}

  The parameters to be trained in our model are the weight matrix $\mathbf{W}$ in the fully connected layer and the vector representations of the words, contextual entities, and target entities.
  The weight matrix was initialized at random and the vector representations were initialized using pre-trained representations.
  The pre-trained representations of words and entities were learned jointly using the skip-gram model \cite{Mikolov2013,Mikolov2013a} with negative sampling\footnote{We used the skip-gram model implemented in the open-source Gensim library with $size = 300$, $window = 10$, $negative = 15$, $min\_count = 3$, and $iter = 5$. Default values were used for other parameters.}.
  The corpus was automatically generated by replacing the name of each entity annotation in the Wikipedia documents with a unique identifier of the entity corresponding to that annotation.
  Note that we used the same pre-trained entity representations to initialize the representations of the contextual entities and the target entities.
  Additionally, we used all Wikipedia documents obtained from the July 2016 version of Wikipedia dump\footnote{We obtained the Wikipedia dump from Wikimedia Downloads: \url{https://dumps.wikimedia.org/}} to build the corpus.

  \subsection{Implementation Details}

  The proposed model was implemented using PyTorch\footnote{\url{http://pytorch.org}} and trained with mini-batch stochastic gradient descent (SGD).
  The mini-batch size was fixed at 100 and the learning rate was automatically controlled by Adadelta \cite{Zeiler2012}.
  We trained the model by iterating over the documents in the KB in random order for 50 epochs\footnote{We experimented using 10, 20, 30, and 50 epochs. All numbers achieved similar performance in our experiments. We used the model trained for 50 epochs because it achieved the best P@1 performance in our fine-grained entity typing task.}.
  For computational efficiency, we used only the first 2,000 words and first 300 entities in the documents.
  The training took approximately 25 h on an NVIDIA GTX 1080 Ti GPU.
  Regarding the other hyper-parameters, the representations were set to have $d = 300$ dimensions, the size of the negative entities was $k = 100$, and the dropout probability was set to $p = 0.5$, as recommended in \newcite{JMLR:v15:srivastava14a}

  \section{Experiments}

  To evaluate the models described in the previous section, we conducted fine-grained entity typing and text classification tasks using the learned representations.
  A description of each task is given in the following subsections.
  Finally, we qualitatively analyze the learned representations.

  \subsection{Fine-grained Entity Typing}
  \label{subsec:entity-typing}

  This section describes the task of fine-grained entity typing \cite{yaghoobzadeh-schutze:2015:EMNLP,neelakantan-chang:2015:NAACL-HLT,yaghoobzadeh-schutze:2017:EACLlong} using the entity representations learned by our proposed models.
  The aim of this task is to assign each entity with one or more fine-grained types such as \textit{musician} and \textit{film}.
  Because an entity typing model is capable of predicting the entity types that are missing from the KB, this can be seen as a \textit{knowledge base completion} problem.
  The task is important because entity type information is often missing from KBs, but is known to be beneficial for various downstream natural language tasks such as entity linking \cite{Ling2015}, coreference resolution \cite{hajishirzi-EtAl:2013:EMNLP}, and semantic parsing \cite{Liu:2015:TWE:2886521.2886657}.

  \subsubsection*{Setup}

  Our experimental setup follows that of \newcite{yaghoobzadeh-schutze:2015:EMNLP}.
  In particular, we use their entity dataset of 201,933 Freebase\footnote{\url{https://developers.google.com/freebase/}} entities mapped to 102 entity types based on the FIGER type set \cite{Ling2012}.
  The dataset consists of a training set (50\%), development set (20\%), and test set (30\%).
  Because the dataset is constructed based on Freebase, we preprocessed the data by mapping each entity to the corresponding entry in Wikipedia and excluded those entities that did not exist in Wikipedia.\footnote{We used the \textit{wikipedia.en\_title} property contained in the Freebase dump to create the mapping.}
  As a result, we successfully mapped approximately 92\% of the entities to Wikipedia, and obtained training, development, and test sets containing 93,350, 37,036, and 55,715 entities, respectively.
  We publicized the dataset and the code used to generate the dataset at \url{https://github.com/studio-ousia/textent/}.

  Following \newcite{yaghoobzadeh-schutze:2015:EMNLP}, we evaluated the models using ranking and classification measures.
  The ranking measures test how well a model ranks entity types.
  In particular, we ranked the entity types based on the probabilities assigned by the model and evaluated the ranked list using the precision at 1 (P@1) and breakeven point (BEP)\footnote{BEP is the F1 score at the point in the ranked list at which the precision and recall have the same value.}.

  The classification measures evaluate the quality of the thresholded assignment decisions of a model.
  The assignment decisions are based on thresholding the probability assigned to each type.
  The threshold is selected per type by maximizing the F1 score of entities assigned to the type in the development set.
  We used the accuracy (an entity is correct if all its types and no incorrect types are assigned to it), micro-average F1 (F1 score of all type--entity assignment decisions), and macro-average F1 (F1 score of types assigned to an entity, averaged over entities).
  These ranking and classification measures are exactly the same as those used in \newcite{yaghoobzadeh-schutze:2015:EMNLP}.

  \subsubsection*{Method}

  We used an MLP classifier with the entity representations as inputs to predict the probability of entity $e$ being a member of type $t$ in the set of possible types $T$.
  In particular, we used an MLP with a single hidden layer and the tanh activation function, and an output layer that contains, for each possible type $t \in T$, a logistic regression classifier that predicts the probability of $t$:
  \begin{equation}
  \bigl[P(t_1|e), ..., P(t_{|T|}|e)\bigr] = \sigma \bigl( \mathbf{W}_o\,tanh \left( \mathbf{W}_{h}\mathbf{c}_e\right) \bigr),
  \end{equation}
  where $\mathbf{c}_e \in \mathbb{R}^d$ is the vector representation of entity $e$, $\sigma$ is the sigmoid function, and $\mathbf{W}_h \in \mathbb{R}^{h \times d}$ and $\mathbf{W}_o \in \mathbb{R}^{|T| \times h}$ are the weight matrices corresponding to the hidden layer and the output layer, respectively.
  The model was trained to minimize the binary cross-entropy loss summed over all entities and types:
  \begin{equation}
  -\sum_e\sum_t\bigl(y_{e,t} \log p_{e,t} + (1 - y_{e,t}) \log (1 - p_{e,t})\bigr),
  \end{equation}
  where $y_{e,t} \in \{0, 1\}$ and $p_{e,t}$ denote the ground-truth label and predicted probability, respectively, of entity $e$ being type $t$.
  The parameters in $\mathbf{W}_h$ and $\mathbf{W}_o$ are updated in the training stage.
  Note that the model described here is equivalent to that proposed in \newcite{yaghoobzadeh-schutze:2015:EMNLP}.

  The model was trained using mini-batch SGD, with the learning rate controlled by Adam \cite{kingma2014adam} and the mini-batch size set to 32.
  The model was trained using the training set and evaluated using the test set.
  Following \newcite{yaghoobzadeh-schutze:2015:EMNLP}, the number of hidden units was set to 200.
  We also measured P@1 on the development set to locate the best epoch for testing.

  \subsubsection*{Baselines}

  The performance of our models is compared with that of the following three entity representation models.

  \begin{itemize}
    \item \textbf{Figment-GM} \cite{yaghoobzadeh-schutze:2015:EMNLP} is based on the skip-gram model \cite{Mikolov2013,Mikolov2013a} trained using a large corpus with automatically generated entity annotations (i.e., FACC1 \cite{Gabrilovich2013}).
    In this experiment, we used the entity representations publicized by the authors\footnote{\url{https://github.com/yyaghoobzadeh/figment}}.
    \item \textbf{Skip-Gram-Wiki} is equivalent to Figment-GM, except that Wikipedia is used as the entity-annotated corpus.
    This model is also the same as our pre-trained representations described in Section \ref{subsec:parameters}.
    \item \textbf{Wikipedia2Vec} \cite{Yamada2016} extends the skip-gram model to learn entity representations based on the contextual words of link anchors in Wikipedia and the internal link structure of Wikipedia entities.
    We used the entity representations trained using the code publicized by the authors\footnote{\url{https://github.com/wikipedia2vec/wikipedia2vec}} and the Wikipedia dump used to train the Skip-Gram-Wiki model.\footnote{We trained the representations with $dim\_size = 300$, $window = 10$, $negative = 15$, $min\_entity\_count = 3$, and $iteration = 5$. Default values were used for other parameters.}
  \end{itemize}

  We used the entity typing method presented above with the entity representations of each baseline model as inputs.
  Note that, because the Wikipedia2Vec and Skip-Gram-Wiki models were trained using the link anchors in Wikipedia, they do not contain entities that do not appear or are very rare as the link anchor destinations in Wikipedia.
  To address this, we evaluated these models in the following two settings:
  (1) using only the entities that exist in the model, and
  (2) using all entities, including non-existent ones.
  In the latter setting,  we used the zero vector as the representation of non-existent entities.
  Similar to the latter setting, the former is not a fair comparison because it is typically more difficult to learn good entity representations of rare entities than those of popular entities \cite{yaghoobzadeh-schutze:2015:EMNLP}.

  \subsubsection*{Results}

  \begin{table}[t]
    \centering
    \begin{tabular}{l|cc|ccc}
                              & P@1           & BEP           & Acc.          & Mic.          & Mac.         \\
      \hline
      TextEnt-full            & \textbf{.932} & \textbf{.948} & \textbf{.626} & \textbf{.857} & \textbf{.842}\\
      TextEnt-word            & .909          & .933          & .611          & .838          & .820         \\
      TextEnt-entity          & .882          & .912          & .560          & .702          & .770         \\
      \hline
      Figment-GM              & .813          & .858          & .421          & .719          & .683         \\
      Wikipedia2Vec           & .925          & .943          & .600          & .844          & .822         \\
      Wikipedia2Vec (all)     & .897          & .917          & .554          & .798          & .787         \\
      Skip-Gram-Wiki          & .900          & .927          & .576          & .831          & .804         \\
      Skip-Gram-Wiki (all)    & .852          & .881          & .510          & .764          & .740         \\
    \end{tabular}
    \caption {Results of the entity typing task.}
    \label{tb:entity-typing-results}
  \end{table}

  Table \ref{tb:entity-typing-results} compares the results of our models with those of the baseline models.
  Our TextEnt-full model outperforms the baseline models in all measures.
  In particular, the TextEnt-full model achieves a strong P@1 score of 93.2\%, which clearly shows the effectiveness of our entity typing model for many downstream NLP tasks.
  Moreover, the TextEnt-full model generally performs better than both the TextEnt-word and TextEnt-entity models.
  This demonstrates the effectiveness of combining the semantic signals obtained from words and entities.

  \subsection{Multiclass Text Classification}
  \label{subsec:text-classification}

  This section describes the multiclass text classification task, which tests the ability of our proposed representations to encode arbitrary documents.
  Our key assumption here is that, because our proposed representations are trained to predict the corresponding entity of a given document in the KB, they can also classify non-KB documents into classes that are much more \textit{coarse-grained} than entities.

  \subsubsection*{Setup}

  Following \newcite{Jin2016}, we used two standard text classification datasets: the 20 newsgroups dataset\footnote{We used the \textit{by-date} version of the dataset obtained from \url{http://qwone.com/~jason/20Newsgroups/}.} (denoted by 20NG) \cite{Lang1995} and the R8 dataset \cite{Debole2005}.
  The 20NG dataset consists of 11,314 training documents and 7,532 test documents retrieved from 20 different newsgroups.
  The documents are partitioned nearly equally across the classes.
  The R8 dataset contains documents from the eight most frequent classes of the Reuters-21578 corpus \cite{Lewis1992}, which consists of labeled news articles from the 1987 Reuters newswire.
  The R8 dataset contains 5,485 documents for training and 2,189 documents for testing.
  Unlike the 20NG dataset, the R8 dataset is imbalanced; the largest class contains 3,923 documents and the smallest class contains 51 documents.
  For both datasets, we report the accuracy and macro-average F1 score.
  Furthermore, the development set was formed by selecting 10\% of the documents in the training set at random for both datasets.

  As preprocessing, we lowercased all words and removed words and entities appearing fewer than five times.
  Furthermore, we automatically annotated entity mentions in the documents using an entity linking system.
  In particular, we used TAGME\footnote{We used the public Web API service available at \url{https://services.d4science.org/}.} \cite{Ferragina2010}, a state-of-the-art entity linking system that is freely available and has been frequently used in recent studies \cite{Xiong2016,Hasibi2016}.
  However, TAGME returned many irrelevant entity mentions that would act as noise (e.g., \textit{I like} refers to an entity \textit{I Like (Keri Hilson song)}).
  Thus, we excluded mentions having relevance scores\footnote{We used the $\rho$ scores assigned by TAGME.} of less than 0.05\footnote{Excluding entity mentions using the relevance scores is the recommended practice described in the documentation: \url{https://services.d4science.org/web/tagme/documentation}}.

  \subsubsection*{Method}

  For this task, we simply stacked a logistic regression layer onto our TextEnt model to classify documents into the predefined classes.
  First, we encoded each document (words with entity annotations) and used the resulting document representation (i.e., $\mathbf{v}_D$ in the TextEnt-full model, $\mathbf{v}_{D_w}$ in the TextEnt-word model, and $\mathbf{v}_{D_e}$ in the TextEnt-entity model) as the feature of the logistic regression classifier.

  We trained the classifier using the training set of each dataset, and evaluated the classification performance using the corresponding test set.
  The classifier was trained using mini-batch SGD, with the learning rate controlled by Adam \cite{kingma2014adam} and the mini-batch size set to 32.
  The accuracy on the development set of each dataset was used to locate the best epoch for testing.

  \subsubsection*{Baselines}

  We adopted the following state-of-the-art models as our baselines.

  \begin{itemize}
    \item \textbf{BoW-SVM} is based on a linear support vector machine (SVM) classifier with bag-of-words (BoW) features as inputs.
    This model outperforms the conventional naive Bayes model \cite{Jin2016}.
    \item \textbf{BoE} \cite{Jin2016} is an extension of the skip-gram model that learns different word representations per target class.
    A linear model based on learned word representations was used to classify documents.
    This model achieves state-of-the-art results on both the 20NG and R8 datasets.
  \end{itemize}
  We also used the \textbf{Wikipedia2Vec} and \textbf{Skip-Gram-Wiki} models described in Section \ref{subsec:entity-typing} as baselines.
  For this experiment, we simply input the representations of words and entities in these models to our text classification model described in the previous section.

  \subsubsection*{Results}

  Table \ref{tb:text-classification-results} compares the results of our proposed models with those of the baseline models.
  We obtained the BoW-SVM and BoE results from \newcite{Jin2016}.
  Our TextEnt-full model outperforms the state-of-the-art models in terms of accuracy and macro F1 score on both the 20NG and R8 datasets.
  Furthermore, similar to the results of our previous experiment, the TextEnt-full model generally performs better than both the TextEnt-word and TextEnt-entity models.
  This shows that combining semantic signals obtained from words and entities is also beneficial for text classification tasks.

  Furthermore, we conducted a detailed comparison of the BoW-SVM model, BoE model, and TextEnt-full model using the class-level F1 scores on the 20NG dataset (Table \ref{tb:20ng-class-level-results}) and the R8 dataset (Table \ref{tb:r8-class-level-results}).
  On the 20NG dataset, our model achieves the best scores in more than half of the classes and provides comparable performance in the other classes.
  Moreover, our model achieves strong performance in classes with relatively few documents on the R8 dataset.
  This is because our model successfully captures the strong semantic signals that can only be obtained from entities.

  \begin{table}[t]
    \centering
    \begin{tabular}{l|cc|cc}
      \multirow{2}{*}{} & \multicolumn{2}{c|}{20NG} & \multicolumn{2}{c}{R8}\\
      & Acc. & F1 & Acc. & F1\\
      \hline
      TextEnt-full      & \textbf{.845} & \textbf{.839} & \textbf{.967} & \textbf{.910} \\
      TextEnt-word      & .836 & .828 & .965 & .860 \\
      TextEnt-entity    & .831 & .824 & .957 & .878 \\
      \hline
      BoW-SVM        & .790 & .783 & .947 & .851 \\
      BoE            & .831 & .827 & .965 & .886 \\
      Wikipedia2Vec  & .829 & .823 & .965 & .881 \\
      Skip-Gram-Wiki & .822 & .815 & .963 & .879 \\
    \end{tabular}
    \caption{Results of the text classification task.}
    \label{tb:text-classification-results}
  \end{table}

  \begin{table}[t]
    \centering
    \setlength\tabcolsep{6pt}
    \begin{tabular}{l|ccc}
      Class & SVM & BoE & TextEnt \\
      \hline
      alt.atheism              &         .699  &         .712  & \textbf{.783} \\
      comp.graphics            &         .702  &         .724  & \textbf{.773} \\
      comp.os.ms-windows.misc  &         .714  &         .724  & \textbf{.742} \\
      comp.sys.ibm.pc.hardware &         .673  &         .706  & \textbf{.721} \\
      comp.sys.mac.hardware    &         .778  &         .792  & \textbf{.840} \\
      comp.windows.x           &         .779  & \textbf{.853} &         .846  \\
      misc.forsale             &         .846  & \textbf{.852} &         .829  \\
      rec.autos                &         .817  & \textbf{.910} &         .909  \\
      rec.motorcycles          &         .900  &         .942  & \textbf{.943} \\
      rec.sport.baseball       &         .895  & \textbf{.947} &         .941  \\
      rec.sport.hockey         &         .935  & \textbf{.967} &         .960  \\
      sci.crypt                &         .890  &         .926  & \textbf{.934} \\
      sci.electronics          &         .721  &         .737  & \textbf{.757} \\
      sci.med                  &         .803  &         .869  & \textbf{.891} \\
      sci.space                &         .892  &         .885  & \textbf{.900} \\
      soc.religion.christian   &         .823  &         .877  & \textbf{.904} \\
      talk.politics.guns       &         .781  & \textbf{.833} &         .810  \\
      talk.politics.mideast    &         .837  &         .920  & \textbf{.944} \\
      talk.politics.misc       & \textbf{.699} &         .687  &         .678  \\
      talk.religion.misc       &         .590  & \textbf{.676} &         .672  \\
    \end{tabular}
    \caption{Class-level F1 scores in each class on the 20NG dataset.}
    \label{tb:20ng-class-level-results}
  \end{table}

  \begin{table}[t]
    \centering
    \setlength\tabcolsep{6pt}
    \begin{tabular}{l|c|ccc}
      Class    & Count  & SVM  & BoE  & TextEnt\\
      \hline
      grain    & 51     & .824 &         .818  & \textbf{.889} \\
      ship     & 144    & .781 &         .783  & \textbf{.829} \\
      interest & 271    & .745 &         .832  & \textbf{.873} \\
      money-fx & 293    & .687 &         .853  & \textbf{.876} \\
      trade    & 326    & .897 &         .879  & \textbf{.918} \\
      crude    & 374    & .929 & \textbf{.958} &         .929 \\
      acq      & 2,292  & .956 & \textbf{.978} &         .977 \\
      earn     & 3,923  & .986 & \textbf{.990} &         .988 \\
    \end{tabular}
    \caption{Class-level F1 scores with the number of documents in each class on the R8 dataset.}
    \label{tb:r8-class-level-results}
  \end{table}

  \section{Qualitative Analysis}

  \begin{table*}[t]
    \centering
    \begin{tabular}{p{1.8cm}|p{5.5cm}|p{7cm}}
      Class & Sentence  & Nearest entities\\
      \hline
      sci.space &
      At one time there was speculation that the first spacewalk (Alexei Leonov?) was a staged fake. &
      Sputnik 1 (0.39),
      Soviet space program (0.38),
      Soyuz 5 (0.38),
      Vostok 1 (0.37)
      \\
      \hline
      rec.autos &
      I prefer a manual to an automatic as it should be. &
      Manual transmission (0.45),
      Automatic transmission (0.45),
      Dual-clutch transmission (0.43),
      Semi-automatic transmission (0.41)
      \\
      \hline
      sci.crypt &
      I change login passwords every couple of months. &
      Password (0.49),
      Login (0.46),
      Privilege (computing) (0.44),
      Privilege escalation (0.43)
      \\
      \hline
      soc.religion. christian &
      Which version of the Bible do you consider to be the most accurate translation? &
      Bible translations (0.37),
      King James Only movement (0.37),
      Biblical poetry (0.36),
      The Living Bible (0.36)
      \\
      \hline
      sci.med &
      The blood tests have shown that I have a little too much Hemoglobin &
      Blood (0.38),
      Introduction to genetics (0.38),
      Hemoglobin (0.37),
      Blood transfusion (0.35)
      \\
    \end{tabular}
    \caption{Five example sentences with their top nearest entities using the TextEnt model.}
    \label{tb:example-sentences}
  \end{table*}
  To investigate how our model encodes documents and entities into the same continuous vector space, we extracted five example sentences from the 20NG dataset and encoded each sentence into a vector using our model. The closest entities to this vector based on the cosine similarity are presented in Table \ref{tb:example-sentences}.
  We automatically annotated the entity mentions using TAGME\footnote{We used the same configuration as described in Section \ref{subsec:text-classification}.}, and fed the words and detected entities into the TextEnt-full model.
  Table \ref{tb:example-sentences} presents the sentences, nearest entities, and their corresponding classes in the 20NG dataset.
  Our model successfully encodes the sentences into vectors that are close to their relevant entities.
  For example, all nearest entities of the first sentence ``\textit{At one time there was speculation that the first spacewalk (Alexei Leonov?) was a staged fake}'' are strongly related to the historic Soviet space program.
  Similar results can be observed in the other four examples.

  \section{Related Work}
  In recent years, various models for computing distributed representations of text (e.g., sentences and documents) have been proposed \cite{DBLP:conf/icml/LeM14,NIPS2015_5950,Wieting2015,TACL711}.
  These models typically use large, unstructured corpora for training; however, certain models attempt to learn text representations from structured data.
  For instance, \newcite{TACL711} proposed a neural network model that learns text representations from online public dictionaries by predicting each dictionary word from its description.
  Further, \newcite{Wieting2015} used a large set of paraphrase pairs obtained from the Paraphrase Database \cite{ganitkevitch-vandurme-callisonburch:2013:NAACL-HLT} to learn text representations.

  A number of recent models have attempted to learn distributed representations of entities from a KB.
  For example, \newcite{hu-EtAl:2015:ACL-IJCNLP} extended the skip-gram model \cite{Mikolov2013} to learn entity representations using the hierarchical structure of the KB, and \newcite{Li2016} modified the model by Hu et al. to learn both the category representations and entity representations using the category information of the KB.
  Additionally, \textit{relational embedding} models \cite{Bordes2013,wang-EtAl:2014:EMNLP20145,AAAI159571} learn the entity representations for link prediction tasks.

  Furthermore, some models learn the representations of both words and entities from the KB.
  A simple method reported in the literature \cite{yaghoobzadeh-schutze:2015:EMNLP,TACL1065} is used to derive the pre-trained representations in this study (i.e., preprocessing an entity-annotated corpus by replacing the name of each annotation with the unique identifier of the entity and feeding the corpus into a word embedding model (e.g., skip-gram)).
  \newcite{Yamada2016} proposed Wikipedia2Vec, which extends this idea by using neighboring entities in the internal link graph of the KB as additional contexts for training the model.
  Note that we used Wikipedia2Vec as a baseline method in the two experiments conducted in this study.
  Similarly, in their subsequent work \cite{TACL1065}, they proposed a neural network model that takes entity-annotated text as input and learns word and entity representations by predicting the annotated entities contained in each text.
  Furthermore, \newcite{mancini-EtAl:2017:CoNLL} proposed a model that maps words and entities in a lexical dictionary (i.e., BabelNet \cite{NAVIGLI2012217}) to a single vector space by extending the CBOW model.
  Unlike our proposed model, these models require users to design a composition function (e.g., vector averaging) to model the semantics of a document using words and entities in it.
  Moreover, we showed that our approach is highly effective for the two important tasks of fine-grained entity typing and multiclass text classification.

  \section{Conclusions}
  In this paper, we described \textit{TextEnt}, a simple neural network model that learns distributed representations of entities and documents from large-scale KB descriptions.
  We evaluated the performance of the proposed model on fine-grained entity typing and text classification tasks, and achieved state-of-the-art results in both cases, which clearly demonstrates the effectiveness of our approach.
  In future work, we will explore the applicability of our model to broader NLP tasks such as entity search and KB-based question answering.

  \section*{Acknowledgements}
  We would like to thank the anonymous reviewers for their careful reading of our manuscript and their helpful comments and suggestions.

  \bibliographystyle{acl}
  \bibliography{library}

\begin{thebibliography}{}

\bibitem[\protect\citename{Bordes \bgroup et al.\egroup }2013]{Bordes2013}
Antoine Bordes, Nicolas Usunier, Alberto Garcia-Duran, Jason Weston, and Oksana
  Yakhnenko.
\newblock 2013.
\newblock {Translating Embeddings for Modeling Multi-relational Data}.
\newblock In {\em Advances in Neural Information Processing Systems 26}, pages
  2787--2795.

\bibitem[\protect\citename{Debole and Sebastiani}2005]{Debole2005}
Franca Debole and Fabrizio Sebastiani.
\newblock 2005.
\newblock {An Analysis of the Relative Hardness of Reuters-21578 Subsets:
  Research Articles}.
\newblock {\em Journal of the American Society for Information Science and
  Technology}, 56(6):584--596.

\bibitem[\protect\citename{Ferragina and Scaiella}2010]{Ferragina2010}
Paolo Ferragina and Ugo Scaiella.
\newblock 2010.
\newblock {TAGME: On-the-fly Annotation of Short Text Fragments (by Wikipedia
  Entities)}.
\newblock In {\em Proceedings of the 19th ACM International Conference on
  Information and Knowledge Management}, pages 1625--1628.

\bibitem[\protect\citename{Gabrilovich \bgroup et al.\egroup
  }2013]{Gabrilovich2013}
Evgeniy Gabrilovich, Michael Ringgaard, and Amarnag Subramanya.
\newblock 2013.
\newblock {FACC1: Freebase Annotation of ClueWeb Corpora, Version 1 (Release
  date 2013-06-26, Format version 1, Correction level 0)}.

\bibitem[\protect\citename{Ganitkevitch \bgroup et al.\egroup
  }2013]{ganitkevitch-vandurme-callisonburch:2013:NAACL-HLT}
Juri Ganitkevitch, Benjamin {Van Durme}, and Chris Callison-Burch.
\newblock 2013.
\newblock {PPDB: The Paraphrase Database}.
\newblock In {\em Proceedings of the 2013 Conference of the North American
  Chapter of the Association for Computational Linguistics: Human Language
  Technologies}, pages 758--764.

\bibitem[\protect\citename{Hajishirzi \bgroup et al.\egroup
  }2013]{hajishirzi-EtAl:2013:EMNLP}
Hannaneh Hajishirzi, Leila Zilles, Daniel~S Weld, and Luke Zettlemoyer.
\newblock 2013.
\newblock {Joint Coreference Resolution and Named-Entity Linking with
  Multi-Pass Sieves}.
\newblock In {\em Proceedings of the 2013 Conference on Empirical Methods in
  Natural Language Processing}, pages 289--299.

\bibitem[\protect\citename{Hasibi \bgroup et al.\egroup }2016]{Hasibi2016}
Faegheh Hasibi, Krisztian Balog, and Svein~Erik Bratsberg.
\newblock 2016.
\newblock {Exploiting Entity Linking in Queries for Entity Retrieval}.
\newblock In {\em Proceedings of the 2016 ACM International Conference on the
  Theory of Information Retrieval}, pages 209--218.

\bibitem[\protect\citename{Hill \bgroup et al.\egroup }2016]{TACL711}
Felix Hill, Kyunghyun Cho, Anna Korhonen, and Yoshua Bengio.
\newblock 2016.
\newblock {Learning to Understand Phrases by Embedding the Dictionary}.
\newblock {\em Transactions of the Association for Computational Linguistics},
  4:17--30.

\bibitem[\protect\citename{Hu \bgroup et al.\egroup
  }2015]{hu-EtAl:2015:ACL-IJCNLP}
Zhiting Hu, Poyao Huang, Yuntian Deng, Yingkai Gao, and Eric Xing.
\newblock 2015.
\newblock {Entity Hierarchy Embedding}.
\newblock In {\em Proceedings of the 53rd Annual Meeting of the Association for
  Computational Linguistics and the 7th International Joint Conference on
  Natural Language Processing (Volume 1: Long Papers)}, pages 1292--1300.

\bibitem[\protect\citename{Iyyer \bgroup et al.\egroup }2015]{iyyer-EtAl:2015}
Mohit Iyyer, Varun Manjunatha, Jordan Boyd-Graber, and Hal {Daum{\'{e}} III}.
\newblock 2015.
\newblock {Deep Unordered Composition Rivals Syntactic Methods for Text
  Classification}.
\newblock In {\em Proceedings of the 53rd Annual Meeting of the Association for
  Computational Linguistics and the 7th International Joint Conference on
  Natural Language Processing (Volume 1: Long Papers)}, pages 1681--1691.

\bibitem[\protect\citename{Jin \bgroup et al.\egroup }2016]{Jin2016}
Peng Jin, Yue Zhang, Xingyuan Chen, and Yunqing Xia.
\newblock 2016.
\newblock {Bag-of-embeddings for Text Classification}.
\newblock In {\em Proceedings of the Twenty-Fifth International Joint
  Conference on Artificial Intelligence}, pages 2824--2830.

\bibitem[\protect\citename{Kingma and Ba}2014]{kingma2014adam}
Diederik Kingma and Jimmy Ba.
\newblock 2014.
\newblock {Adam: A Method for Stochastic Optimization}.
\newblock {\em arXiv preprint arXiv:1412.6980v9}.

\bibitem[\protect\citename{Kiros \bgroup et al.\egroup }2015]{NIPS2015_5950}
Ryan Kiros, Yukun Zhu, Ruslan~R Salakhutdinov, Richard Zemel, Raquel Urtasun,
  Antonio Torralba, and Sanja Fidler.
\newblock 2015.
\newblock {Skip-Thought Vectors}.
\newblock In {\em Advances in Neural Information Processing Systems 28}, pages
  3294--3302.

\bibitem[\protect\citename{Lang}1995]{Lang1995}
Ken Lang.
\newblock 1995.
\newblock {NewsWeeder: Learning to Filter Netnews}.
\newblock {\em Proceedings of the 12th International Conference on Machine
  Learning}, pages 331--339.

\bibitem[\protect\citename{Le and Mikolov}2014]{DBLP:conf/icml/LeM14}
Quoc~V Le and Tomas Mikolov.
\newblock 2014.
\newblock {Distributed Representations of Sentences and Documents}.
\newblock In {\em Proceedings of the 31th International Conference on Machine
  Learning}, volume~32, pages 1188--1196.

\bibitem[\protect\citename{Lewis}1992]{Lewis1992}
David~D. Lewis.
\newblock 1992.
\newblock {An Evaluation of Phrasal and Clustered Representations on a Text
  Categorization Task}.
\newblock In {\em Proceedings of the 15th Annual International ACM SIGIR
  Conference on Research and Development in Information Retrieval}, pages
  37--50.

\bibitem[\protect\citename{Li \bgroup et al.\egroup }2016]{Li2016}
Yuezhang Li, Ronghuo Zheng, Tian Tian, Zhiting Hu, Rahul Iyer, and Katia
  Sycara.
\newblock 2016.
\newblock {Joint Embedding of Hierarchical Categories and Entities for Concept
  Categorization and Dataless Classification}.
\newblock In {\em Proceedings of the 26th International Conference on
  Computational Linguistics}, pages 2678--2688.

\bibitem[\protect\citename{Lin \bgroup et al.\egroup }2015]{AAAI159571}
Yankai Lin, Zhiyuan Liu, Maosong Sun, Yang Liu, and Xuan Zhu.
\newblock 2015.
\newblock {Learning Entity and Relation Embeddings for Knowledge Graph
  Completion}.
\newblock In {\em Proceedings of the 29th AAAI Conference on Artificial
  Intelligence}, pages 2181--2187.

\bibitem[\protect\citename{Ling and Weld}2012]{Ling2012}
Xiao Ling and Daniel~S. Weld.
\newblock 2012.
\newblock {Fine-Grained Entity Recognition}.
\newblock In {\em Proceedings of the Twenty-Sixth AAAI Conference on Artificial
  Intelligence}, pages 94--100.

\bibitem[\protect\citename{Ling \bgroup et al.\egroup }2015]{Ling2015}
Xiao Ling, Sameer Singh, and Daniel~S. Weld.
\newblock 2015.
\newblock {Design Challenges for Entity Linking}.
\newblock {\em Transactions of the Association for Computational Linguistics},
  3:315--328.

\bibitem[\protect\citename{Liu \bgroup et al.\egroup
  }2015]{Liu:2015:TWE:2886521.2886657}
Yang Liu, Zhiyuan Liu, Tat-Seng Chua, and Maosong Sun.
\newblock 2015.
\newblock {Topical Word Embeddings}.
\newblock In {\em Proceedings of the Twenty-Ninth AAAI Conference on Artificial
  Intelligence}, pages 2418--2424.

\bibitem[\protect\citename{Mancini \bgroup et al.\egroup
  }2017]{mancini-EtAl:2017:CoNLL}
Massimiliano Mancini, Jose Camacho-Collados, Ignacio Iacobacci, and Roberto
  Navigli.
\newblock 2017.
\newblock {Embedding Words and Senses Together via Joint Knowledge-Enhanced
  Training}.
\newblock In {\em Proceedings of the 21st Conference on Computational Natural
  Language Learning}, pages 100--111.

\bibitem[\protect\citename{Mikolov \bgroup et al.\egroup }2013a]{Mikolov2013}
Tomas Mikolov, Greg Corrado, Kai Chen, and Jeffrey Dean.
\newblock 2013a.
\newblock {Efficient Estimation of Word Representations in Vector Space}.
\newblock In {\em Proceedings of the 2013 International Conference on Learning
  Representations}, pages 1--12.

\bibitem[\protect\citename{Mikolov \bgroup et al.\egroup }2013b]{Mikolov2013a}
Tomas Mikolov, Ilya Sutskever, Kai Chen, Greg~S. Corrado, and Jeff Dean.
\newblock 2013b.
\newblock {Distributed Representations of Words and Phrases and their
  Compositionality}.
\newblock In {\em Advances in Neural Information Processing Systems 26}, pages
  3111--3119.

\bibitem[\protect\citename{Navigli and Ponzetto}2012]{NAVIGLI2012217}
Roberto Navigli and Simone~Paolo Ponzetto.
\newblock 2012.
\newblock {BabelNet: The Automatic Construction, Evaluation and Application of
  a Wide-Coverage Multilingual Semantic Network}.
\newblock {\em Artificial Intelligence}, 193(Supplement C):217--250.

\bibitem[\protect\citename{Neelakantan and
  Chang}2015]{neelakantan-chang:2015:NAACL-HLT}
Arvind Neelakantan and Ming-Wei Chang.
\newblock 2015.
\newblock {Inferring Missing Entity Type Instances for Knowledge Base
  Completion: New Dataset and Methods}.
\newblock In {\em Proceedings of the 2015 Conference of the North American
  Chapter of the Association for Computational Linguistics: Human Language
  Technologies}, pages 515--525.

\bibitem[\protect\citename{Srivastava \bgroup et al.\egroup
  }2014]{JMLR:v15:srivastava14a}
Nitish Srivastava, Geoffrey Hinton, Alex Krizhevsky, Ilya Sutskever, and Ruslan
  Salakhutdinov.
\newblock 2014.
\newblock {Dropout: A Simple Way to Prevent Neural Networks from Overfitting}.
\newblock {\em Journal of Machine Learning Research}, 15:1929--1958.

\bibitem[\protect\citename{Wang \bgroup et al.\egroup
  }2014]{wang-EtAl:2014:EMNLP20145}
Zhen Wang, Jianwen Zhang, Jianlin Feng, and Zheng Chen.
\newblock 2014.
\newblock {Knowledge Graph and Text Jointly Embedding}.
\newblock In {\em Proceedings of the 2014 Conference on Empirical Methods in
  Natural Language Processing}, pages 1591--1601.

\bibitem[\protect\citename{Wieting \bgroup et al.\egroup }2016]{Wieting2015}
John Wieting, Mohit Bansal, Kevin Gimpel, and Karen Livescu.
\newblock 2016.
\newblock {Towards Universal Paraphrastic Sentence Embeddings}.
\newblock {\em Proceedings of the 2016 International Conference on Learning
  Representations}.

\bibitem[\protect\citename{Xiong \bgroup et al.\egroup }2016]{Xiong2016}
Chenyan Xiong, Jamie Callan, and Tie-Yan Liu.
\newblock 2016.
\newblock {Bag-of-Entities Representation for Ranking}.
\newblock In {\em Proceedings of the 2016 ACM International Conference on the
  Theory of Information Retrieval}, pages 181--184.

\bibitem[\protect\citename{Yaghoobzadeh and
  Schutze}2015]{yaghoobzadeh-schutze:2015:EMNLP}
Yadollah Yaghoobzadeh and Hinrich Schutze.
\newblock 2015.
\newblock {Corpus-level Fine-grained Entity Typing Using Contextual
  Information}.
\newblock In {\em Proceedings of the 2015 Conference on Empirical Methods in
  Natural Language Processing}, pages 715--725.

\bibitem[\protect\citename{Yaghoobzadeh and
  Sch{\"{u}}tze}2017]{yaghoobzadeh-schutze:2017:EACLlong}
Yadollah Yaghoobzadeh and Hinrich Sch{\"{u}}tze.
\newblock 2017.
\newblock {Multi-level Representations for Fine-Grained Typing of Knowledge
  Base Entities}.
\newblock In {\em Proceedings of the 15th Conference of the European Chapter of
  the Association for Computational Linguistics: Volume 1, Long Papers}, pages
  578--589.

\bibitem[\protect\citename{Yamada \bgroup et al.\egroup }2016]{Yamada2016}
Ikuya Yamada, Hiroyuki Shindo, Hideaki Takeda, and Yoshiyasu Takefuji.
\newblock 2016.
\newblock {Joint Learning of the Embedding of Words and Entities for Named
  Entity Disambiguation}.
\newblock In {\em Proceedings of the 20th SIGNLL Conference on Computational
  Natural Language Learning}, pages 250--259.

\bibitem[\protect\citename{Yamada \bgroup et al.\egroup }2017]{TACL1065}
Ikuya Yamada, Hiroyuki Shindo, Hideaki Takeda, and Yoshiyasu Takefuji.
\newblock 2017.
\newblock {Learning Distributed Representations of Texts and Entities from
  Knowledge Base}.
\newblock {\em Transactions of the Association for Computational Linguistics},
  5:397--411.

\bibitem[\protect\citename{Zeiler}2012]{Zeiler2012}
Matthew~D. Zeiler.
\newblock 2012.
\newblock {ADADELTA: An Adaptive Learning Rate Method}.
\newblock {\em arXiv preprint arXiv:1212.5701v1}.

\end{thebibliography}
\end{document}